\documentclass{article}
\usepackage[utf8]{inputenc}
\usepackage{authblk}
\usepackage{setspace}
\usepackage[margin=1.25in]{geometry}
\usepackage{graphicx}
\graphicspath{ {./figures/} }
\usepackage{subcaption}
\usepackage{amsmath}
\usepackage{lineno}
\usepackage{todonotes}
\usepackage{amssymb}
\usepackage{url} 
\usepackage{hyperref}

\usepackage[style=nejm, 
citestyle=numeric-comp,
sorting=none]{biblatex}
\title{FashionFlow: Leveraging Diffusion Models for Dynamic Fashion Video Synthesis from Static Imagery}

\author[1]{Tasin Islam}
\author[1]{Alina Miron}
\author[1]{XiaoHui Liu}
\author[1]{Yongmin Li}

\affil[1]{Department of Computer Science, Brunel University London, London, UK.}

\date{}

\DeclareUnicodeCharacter{0301}{\'{e}}

\begin{document}
\maketitle

\begin{abstract}
Our study introduces a new image-to-video generator called FashionFlow to generate fashion videos. By utilising a diffusion model, we are able to create short videos from still fashion images. Our approach involves developing and connecting relevant components with the diffusion model, which results in the creation of high-fidelity videos that are aligned with the conditional image. The components include the use of pseudo-3D convolutional layers to generate videos efficiently. VAE and CLIP encoders capture vital characteristics from still images to condition the diffusion model at a global level. Our research demonstrates a successful synthesis of fashion videos featuring models posing from various angles, showcasing the fit and appearance of the garment. Our findings hold great promise for improving and enhancing the shopping experience for the online fashion industry.
\end{abstract} 

\section{Introduction}

The rise of online clothing shopping has brought about challenges for both customers and businesses. Customers often have to guess whether a product will look good on them or fit properly, leading to a higher risk of returns and lower satisfaction for both parties \cite{pachoulakis2012augmented}. Additionally, customers are unable to physically experiment with fashion products, such as feeling and flow when worn, which can result in a less satisfying online shopping experience compared to in-person shopping.  

Currently, there is active ongoing research on how deep learning can help fashion businesses prosper and allow customers to have a better shopping experience \cite{cheng2021fashion}. One area of focus is image-based virtual try-on, which uses a deep learning model to combine images of the customer and the desired clothing product to create a try-on image \cite{han2018viton, wang2018toward}. This helps customers visualise how the clothing item will look on them, making it easier for them to make a purchase decision. Not only can deep learning approaches improve customer shopping experience, but they can make business processes more effective. There are deep learning models that can transform rough fashion design sketches into real products \cite{xian2018texturegan}, allowing businesses to quickly develop and create new products. This speeds up the experimentation process and enables businesses to bring their initial ideas to life in a timely manner.

We have developed a new deep learning framework that can help businesses market their fashion products in a more engaging way. This model creates a short video from a still image, showcasing how a piece of clothing moves and flows when worn by an actor. Using a diffusion model, as well as additional components such as variational autoencoder (VAE) and contrastive language–image pre-training (CLIP) encoder, we are able to capture vital characteristics from the still image and generate a high-fidelity video. The contribution from our work is the following: 
\begin{itemize}
    \item Developed a generative model that produces spontaneous fashion videos with realistic movements from a still image. The model is a latent diffusion model, utilising cross-attention mechanism to combine the noisy latent with the conditioning image. 
    \item Utilised a pre-trained VAE decoder to process each frame from the denoised latent space to synthesise a complete fashion video.
    \item Demonstration of local and global conditioning enables the model to preserve most detail from the conditioning image.
\end{itemize} 

We share our source code and provide pre-trained models on our GitHub repository located at \\\href{https://github.com/1702609/FashionFlow/}{github.com/1702609/FashionFlow}.
 
\section{Background}
\label{background}

In this section, we will discuss the existing literature that shows how deep learning can benefit the fashion industry, including its functionality and potential business and customer benefits. We will be concentrating on generative models that can synthesise images and videos.

\subsection{Generative Adversarial Networks}

Generative Adversarial Networks (GANs) have emerged as cutting-edge technology for image synthesis and generation, with significant advancements made in recent years. Particularly, StyleGANs have showcased the potential of GANs in generating remarkably realistic photographic images \cite{karras2019style, karras2020analyzing}. GANs comprise two neural networks that operate in an adversarial manner, allowing the generator network to produce samples that mimic the underlying dataset while the discriminator network evaluates whether the sample is real or generated \cite{goodfellow2014generative}.

Conditional Generative Adversarial Networks (cGANs) represent a significant extension to GANs, enabling the neural network to take in conditional data and use it to influence the generated outcomes \cite{mirza2014conditional}. This capability makes cGANs a valuable tool in virtual try-on use cases, where the network can leverage conditioning data to generate realistic images of try-on.

There are many models of GANs that can synthesise videos. Some of these models generate videos from pure noise \cite{tulyakov2018mocogan, vondrick2016generating, skorokhodov2022stylegan} while others use conditioning data \cite{li2019storygan, chen2019mocycle, bansal2018recycle, wang2020imaginator}.

MoCoGAN generates unconditioned videos from noise vectors \cite{tulyakov2018mocogan}. The model uses two principles to generate a video - content vector and motion vector. The content vector specifies the object and appearance of the video, while the motion vector specifies the motion and movement of the video. The framework generates a video by mapping a sequence of random vectors to a sequence of video frames. Each random vector consists of a content part and a motion part. The content part remains fixed, while the motion part is a stochastic process. MoCoGAN has two discriminators, one that differentiates between real and fake images and the other that differentiates between real and fake video stacks. This ensures that the generated video is of good quality and has realistic dynamics.

Vondrick et al. utilised a spatiotemporal convolutional architecture to synthesise a video \cite{vondrick2016generating}. The model takes a low-dimensional latent code produced by Gaussian noise as its input. The model comprises two streams. The first stream uses spatiotemporal convolutions to upsample the latent code and generate high-dimensional videos with numerous frames. The second stream controls the background and foreground objects to create the impression of a stationary camera with only the object in motion.

The StyleGAN-V is built on the modified StyleGAN2 architecture \cite{karras2020training} to synthesise videos \cite{skorokhodov2022stylegan}. This model utilises a similar concept as the MoCoGAN \cite{tulyakov2018mocogan}, where the noise vector is split into two vectors - the content vector and the motion vector. However, the motion code in StyleGAN-V is continuous, and this enables the production of very long videos with consistent frames. Unlike previous approaches, where the features of image and video are evaluated separately by multiple discriminators \cite{li2019storygan, tulyakov2018mocogan, wang2020imaginator}, StyleGAN-V employs a single discriminator that is conditioned on the time distances between the frames. This discriminator is more efficient than traditional video discriminators and provides more effective feedback to the generator. The experimental results have shown that the StyleGAN-V model can produce videos for as long as one hour without any issues, which was a significant challenge for its competitors.

There are video GANs that use conditioning data to produce desired videos. For instance, StoryGAN uses a paragraph of text to visualise a story by generating a sequence of images that complement the text \cite{li2019storygan}. The generator does this in two steps. First, a story encoder maps the text into a low-dimensional vector using a multi-layer perceptron (MLP). Second, a context encoder, which is a recurrent neural network (RNN), tracks the story flow and generates the next image to keep the story moving forward. There are two discriminators - an image discriminator and a story discriminator - that evaluate the genuineness of the story and image. The image discriminator measures whether the generated image matches the sentence, while the story discriminator ensures that the generated image sequence aligns with the story.

Mocycle-GAN can translate videos from one form to another, for example, converting videos into segmentation labels and vice versa \cite{chen2019mocycle}. It can even transform day-time videos into sunset or night-time environments. Mocycle-GAN maintains consistency in the video by using optical flow and temporal constraints. The model ensures that the video is smooth by maintaining a similar optical flow between adjacent frames throughout the video. Furthermore, the researchers utilise a technique to ensure that the motion in the video aligns with reality. This involves transferring motion information across different types of videos.

ImaGINator is a model that can generate a video using just one image and a class label to control facial expression and action \cite{wang2020imaginator}. The generator consists of an image encoder that encodes the conditioning image into a latent vector, which is then combined with noise and the class label. The decoder is a more complex component that utilises pseudo-3D convolutional layers. These layers separate the convolutional filters into temporal (1D) and spatial (2D) components. This approach is better because pseudo-3D convolutional layers are more efficient to optimise than the standard 3D convolutional \cite{tran2018closer}. Additionally, the decoder employs a fusion mechanism to maintain the appearance of the video throughout. 

\subsection{Diffusion Model}

GANs and diffusion models are two different techniques used to generate data. While GANs use adversarial training to improve the generator, diffusion models gradually add noise to real data to create a series of distorted versions. The model then learns to reverse this process by removing the noise sequentially to generate a coherent image or other data type, starting from random noise.

There are three main formulations of diffusion models \cite{croitoru2023diffusion}, namely the denoising diffusion probabilistic model (DDPM) \cite{ho2020denoising}, noise conditioned score network (NCSN) \cite{song2019generative}, and stochastic differential equation (SDE) \cite{song2020score}. Essentially, each model incorporates two Markov chains. The forward process transforms data distribution into gradual Gaussian noise, while the reverse process utilises deep neural networks to progressively reverse the Gaussian noise and generate an image or video. Researchers have shown that diffusion models perform better than GANs in generating images \cite{dhariwal2021diffusion}. The study found that diffusion models achieved the best FID score in tasks such as generating images of bedrooms, horses, and cats.

Similar to GANs, diffusion models offer the capability to manipulate the generated images using textual descriptions \cite{rombach2022high} or input images \cite{saharia2022palette}. This adaptability opens up a wide array of possibilities for their use in the fashion industry. Diffusion models have emerged as a promising approach to generating videos. Numerous research studies have investigated the potential use of these models in this domain. Recent works such as \cite{singer2022make, wu2023tune, ho2022imagen, luo2023videofusion, villegas2022phenaki, blattmann2023align, ge2023preserve, girdhar2023emu} have shown that diffusion models are currently the most active area of research in the development of generative models for videos. Notably, these models have demonstrated their ability to generate videos with the best quality and temporal consistency. 

Many researchers have pointed out that training diffusion models with large-scale video datasets are computationally expensive \cite{wu2023tune, ho2022imagen, singer2022make, esser2023structure}. Some techniques have tackled this problem by leveraging models that are trained on datasets consisting of pairs of text and images \cite{singer2022make, wu2023tune}. These models are designed to understand how the world looks based on textual input, and they are then trained on unsupervised video footage to learn how images move in the real world  \cite{singer2022make}. 

In the field of video generation using diffusion models, most models rely on a text-to-video approach, where textual inputs are used to create videos \cite{singer2022make, zhou2022magicvideo, wu2023tune, blattmann2023stable, girdhar2023emu}. Only a few models have demonstrated synthesing videos from conditioning images \cite{singer2022make, wang2024magicvideo, karras2023dreampose}.

The video diffusion model \cite{ho2022video} is the first approach to use the diffusion model for generating unconditional videos. This model generates low-resolution video frames first and then leverages a super-resolution module to upsample them. The key feature of this model is that it is trained to approximate the complex distributions of raw videos, which is computationally challenging. 

Make-A-Video \cite{singer2022make} utilises existing text-to-image models and converts them into text-to-video models through the use of spatiotemporal attention and convolution. These techniques enable the model to preserve the video's quality and ensure it has temporal smoothness. Similar to ImaGINator \cite{wang2020imaginator}, Make-A-Video uses pseudo-3D convolutional layers where the 2D convolution is stacked against a 1D convolution. They would use unsupervised learning on unlabeled video data to teach a model how an image would move. The decoder generates low-resolution, low-framerate video frames, which are then interpolated to a higher frame rate and resolution as data passes through the decoder layers.

MagicVideo is an efficient model for text-to-video synthesis. It uses a latent diffusion model (LDM) \cite{rombach2022high}, which is an efficient approach to denoise the latent space to a lower dimension, making the entire process faster. Compared to other video synthesis tools, such as Make-A-Video \cite{singer2022make}, MagicVideo employs a 2D convolution with temporal computation operators to model both the spatial and temporal features of the video. The temporal computation operator is a lightweight adapter that can effectively exploit the correlation between video frames. 

Tune-a-video trains a text-to-video generator using a single text-video pair and a pre-trained text-to-image model \cite{wu2023tune}. The spatiotemporal attention queries relevant positions in previous frames to ensure consistency with the generated video's temporal dimension. During the inference stage, structure guidance from the source video is incorporated. This means that the latent noise of the source video is obtained from the diffusion model with no textual condition. The noise serves as the starting point for DDIM sampling, which is guided by an edited prompt to preserve the video's motion and structure while basing the content on the prompt.

Like Mocycle-GAN \cite{chen2019mocycle}, there are diffusion models that perform video-to-video translation diffusion model \cite{esser2023structure, karras2023dreampose}. For example, Esser et al. \cite{esser2023structure} encodes the input video to extract its structure, including shapes, object locations, and changes over time, as well as its content, like colours, styles, and lighting. They use this information to control and influence the synthesised video. The resulting video must have the same structure as the input video, but the content can be changed through cross-attention, which adjusts the colour and appearance of the video. This allows the input video to be translated into a different style while respecting its motion and structure.

Given their success in generating high-quality videos, our proposed model also utilises a diffusion model to create fashion videos. The contribution of our work is the design of the diffusion model that generates high-fidelity video from conditional images. Our model captures the appearance of conditional images and synthesises believable movements. It differs from previous video diffusion models as most focus on text-to-video synthesis \cite{singer2022make, zhou2022magicvideo, wu2023tune, blattmann2023stable, girdhar2023emu}.

\subsection{Deep Learning for Fashion Application}

There are several deep learning models that can be used to support the fashion industry, as mentioned in \cite{cheng2021fashion}. For instance, virtual try-ons \cite{han2018viton, wang2018toward} give customers the ability to merge images of clothing items with their own images, allowing them to see how the garment will look and fit on them realistically. Facial makeup transfer allows the model to transfer makeup from a reference image to a source image \cite{chen2019beautyglow}. Additionally, pose transfer can show different viewing angles of fashion products by altering the posture of a person in the image \cite{ma2017pose, dong2019towards}.

Image-based fashion recommendation systems (FRSs) offer consumers a highly personalised shopping experience by leveraging their browsing history and previous purchase records to provide tailored recommendations. Among these FRSs, some employ deep learning techniques \cite{polania2019learning,mcauley2015image}. These advanced systems go beyond basic recommendations by predicting compatibility scores between clothing items, particularly in terms of their style, such as colours and patterns.

For instance, when a customer is interested in purchasing a t-shirt, these models can thoroughly analyse the chosen t-shirt and, based on its style attributes, suggest the most suitable trousers that complement the selected t-shirt. Essentially, they make informed decisions on behalf of the customer, ensuring that the clothing combinations harmonise seamlessly. This functionality greatly improves the shopping experience for customers, making it not only convenient but also enhancing their overall satisfaction with the process \cite{chakraborty2021fashion}. 

Video virtual try-on, as demonstrated in works like \cite{dong2019fw, kuppa2021shineon, zhong2021mv, jiang2022clothformer}, take virtual try-on experiences to the next level by seamlessly integrating clothing onto a person in a video. This dynamic approach enables customers to witness clothing items in motion as they respond and adapt to the wearer's movements. Notably, these models capture the subtle nuances of how clothing products move and flow as the person walks, gestures, or performs various activities. This level of detail and realism is exceptionally informative, as it allows customers to gauge not just how a garment looks when stationary but also how it behaves during real-world activities. Consequently, a virtual video try-on elevates the online shopping experience by providing a holistic view of a clothing item's fit, comfort, and aesthetic appeal in motion.

Although the use of diffusion models in fashion-related tasks is a recent development, only a few studies have explored this approach thus far \cite{karras2023dreampose, cao2023difffashion, bhunia2023person}. Some of these studies include DreamPose \cite{karras2023dreampose}, a fashion video synthesis model that uses conditional poses to guide video synthesis, and DiffFashion \cite{cao2023difffashion}, a model that combines texture from one image with a fashion product. These models demonstrate the applicability of diffusion models in the fashion industry and their potential to revolutionise various aspects of it, from creative content generation and virtual try-ons to personalised styling recommendations and trend forecasting.

Our work solves a similar problem presented in DreamPose \cite{karras2023dreampose}, where we generate videos from a single image. However, we differ from DreamPose in terms of producing spontaneous yet believable motions, while DreamPose uses pre-recorded posture data to guide video synthesis. Additionally, we utilise cross-attention \cite{vaswani2017attention} mechanisms that simultaneously condition all video frames. In contrast, DreamPose synthesises each video frame separately and applies cross-attention individually, making their model much slower than ours. 

Our objective is to continue in this line of work to showcase the immense potential of deep learning in revolutionising the fashion industry. By harnessing diffusion models, we are able to generate high-fidelity fashion videos from conditional images that can effectively showcase the unique features of different fashion products in a captivating manner. This provides businesses with an innovative and enjoyable way to market their fashion products while also exploring new avenues to enhance the overall customer experience. 

\section{Method}

\begin{figure*}[!t]
    \centering
    \includegraphics[width=\textwidth]{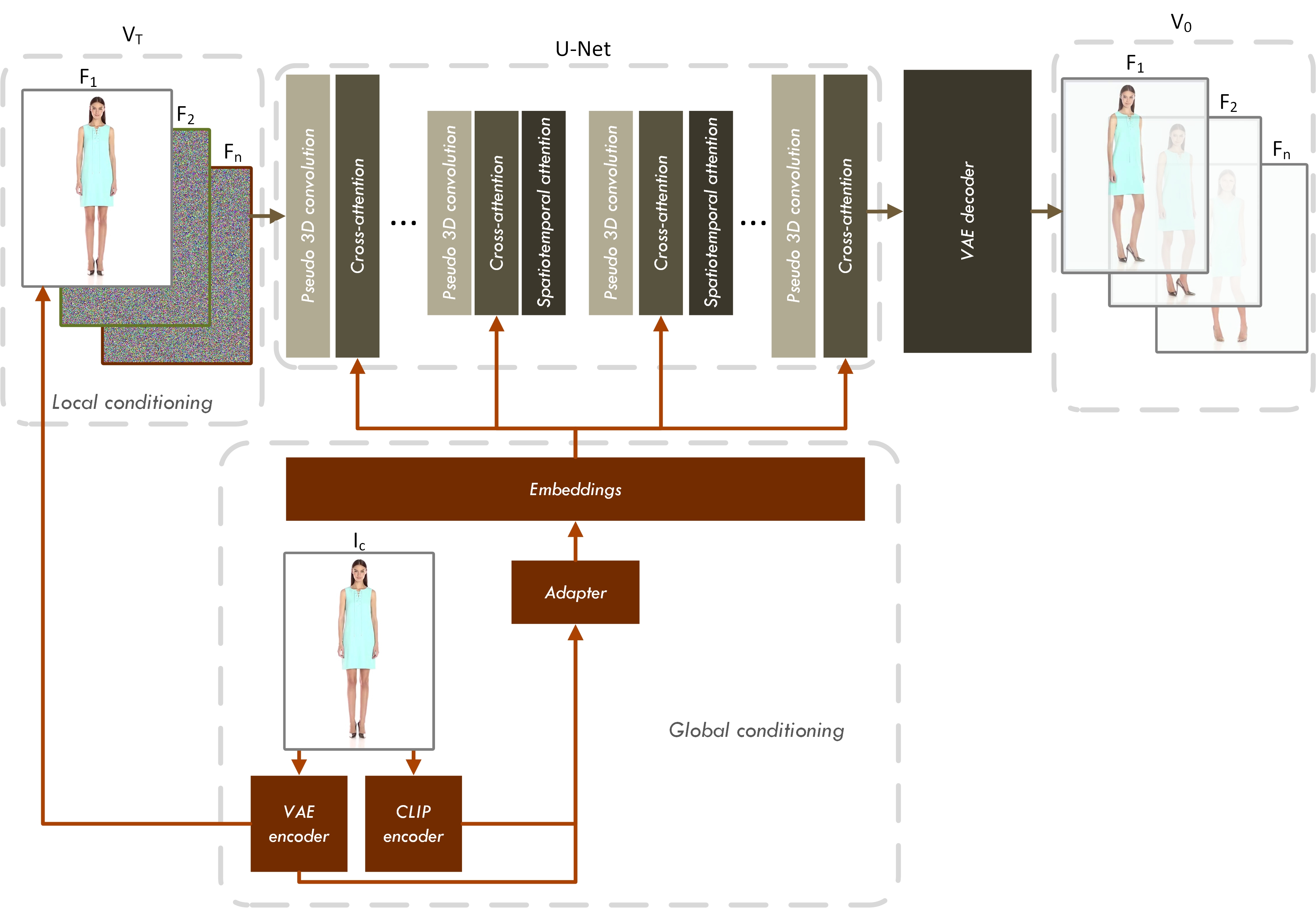}
    \caption{The architecture of our proposed image-to-video model. Our approach involves a latent diffusion model \cite{rombach2022high} to denoise the latent space of a video. Each frame of the latent space is then processed by a pre-trained VAE decoder to generate the final video. We condition the video in two ways: locally and globally. Local conditioning involves adding a VAE-encoded image as the first frame of the noisy latent, while global conditioning involves using cross-attention layers to influence intermediate features with the conditioning image throughout the layers of the U-Net.}
    \label{fig:architecture}
\end{figure*}

In this section, we present an approach for generating short, spontaneous fashion videos using a single image as a conditioner, as shown in Fig.~\ref{fig:architecture}. Our method harnesses the power of the diffusion model. The model produces a video showcasing an actor performing spontaneous yet fashion-relevant poses and movements while maintaining stylistic consistency.

\subsection{Diffusion model}

Using traditional diffusion models to denoise the pixels of videos would be extremely time-consuming and will require significant computational resources. We employ the latent diffusion model (LDM) \cite{rombach2022high} to create a video. LDM directly denoises the latent space, making them more efficient as they require fewer parameters and demand lesser computational resources, thus making them suitable for video synthesis. Generating videos involves producing multiple frames, which can be time-consuming. Therefore, efficiency is a crucial factor, and LDM's direct work on the latent space makes it an ideal option for video synthesis.

There are several ways in which diffusion models add noise to data \cite{croitoru2023diffusion}, and we use the noise scheduler used by the diffusion probabilistic model (DDPM) \cite{ho2020denoising}. The forward process can be described as: 

\begin{equation}
\label{eq:forward_process}
x_t = \sqrt{1-\beta_t} \cdot x_{t-1} + \sqrt{\beta_t} \cdot \epsilon_t, \\
\epsilon_t \sim \mathcal{N}(0, I)
\end{equation}

where at each timestep $t$, the noisy data $x_t$ is produced by adding noise to data from the previous timestep $x_{t-1}$. The noise level at timestep $t$ is denoted by $\beta_t$. Gaussian noise is added to the data at each timestep, which is represented by $\epsilon_t$. Similarly, the reverse process is presented as follows:

\begin{equation}
\label{eq:reverse_process}
x_{t-1} = \mu_\theta(x_t, t) + \sqrt{\beta_t} \cdot \epsilon_t, \\
\epsilon_t \sim \mathcal{N}(0, I)
\end{equation}

where, a neural network $\mu_\theta(x_t, t)$ is used to predict the noise of $x_t$ and subtract it to get $x_{t-1}$.

In equation \ref{eq:forward_process} and \ref{eq:reverse_process}, the variable $x$ is replaced with the latent video $V$, as shown in Figure \ref{fig:architecture}. The video latent space $V_T$ contains complete noise, except for the first frame, which includes a VAE-encoded condition image $I_{VAE}$. We train the U-Net not to modify the first frame by not adding any noise to it during the forward process. Additionally, we use global conditioning to ensure that the video retains high fidelity while preserving the details from $I_c$.

\subsection{Pseudo-3D convolution}

\begin{figure*}[!t]
    \centering
    \begin{subfigure}[t]{0.26\textwidth}   
        \centering
        \includegraphics[width=\columnwidth]{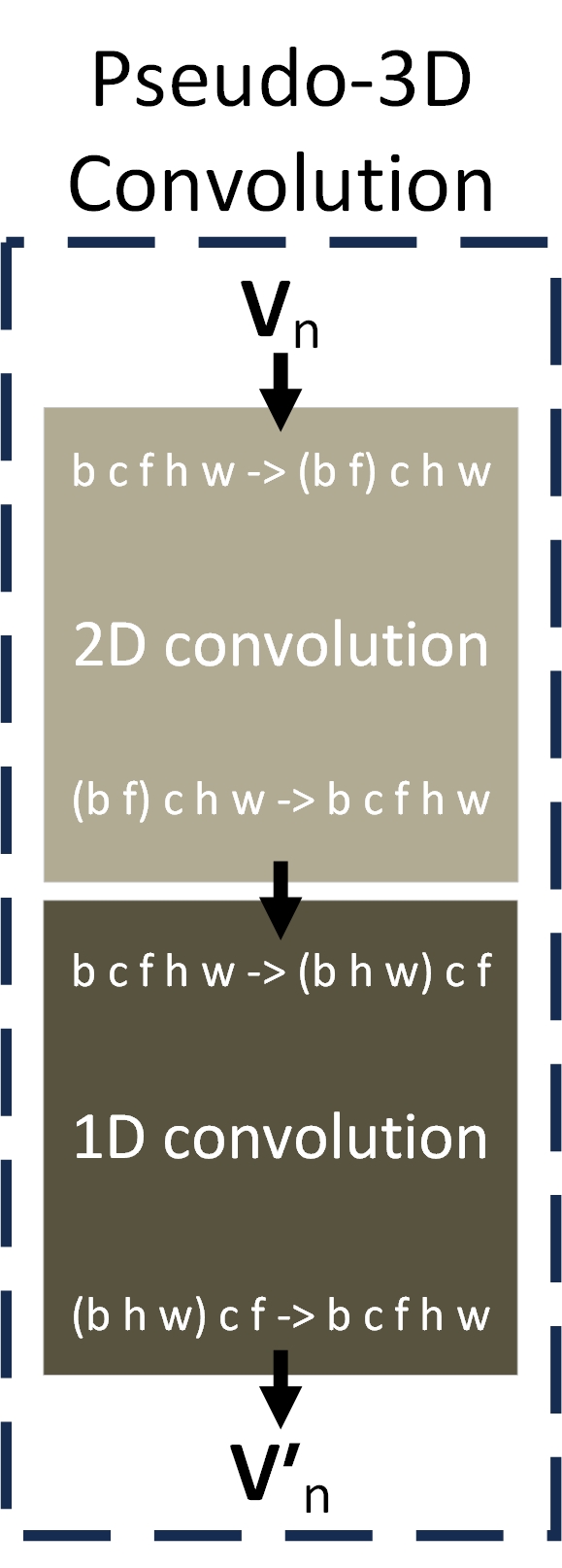}
        \caption[]%
        {} 
        \label{fig:pseudo3D}
    \end{subfigure}
    \hfill
    \begin{subfigure}[t]{0.305\textwidth} 
        \centering
        \includegraphics[width=\columnwidth]{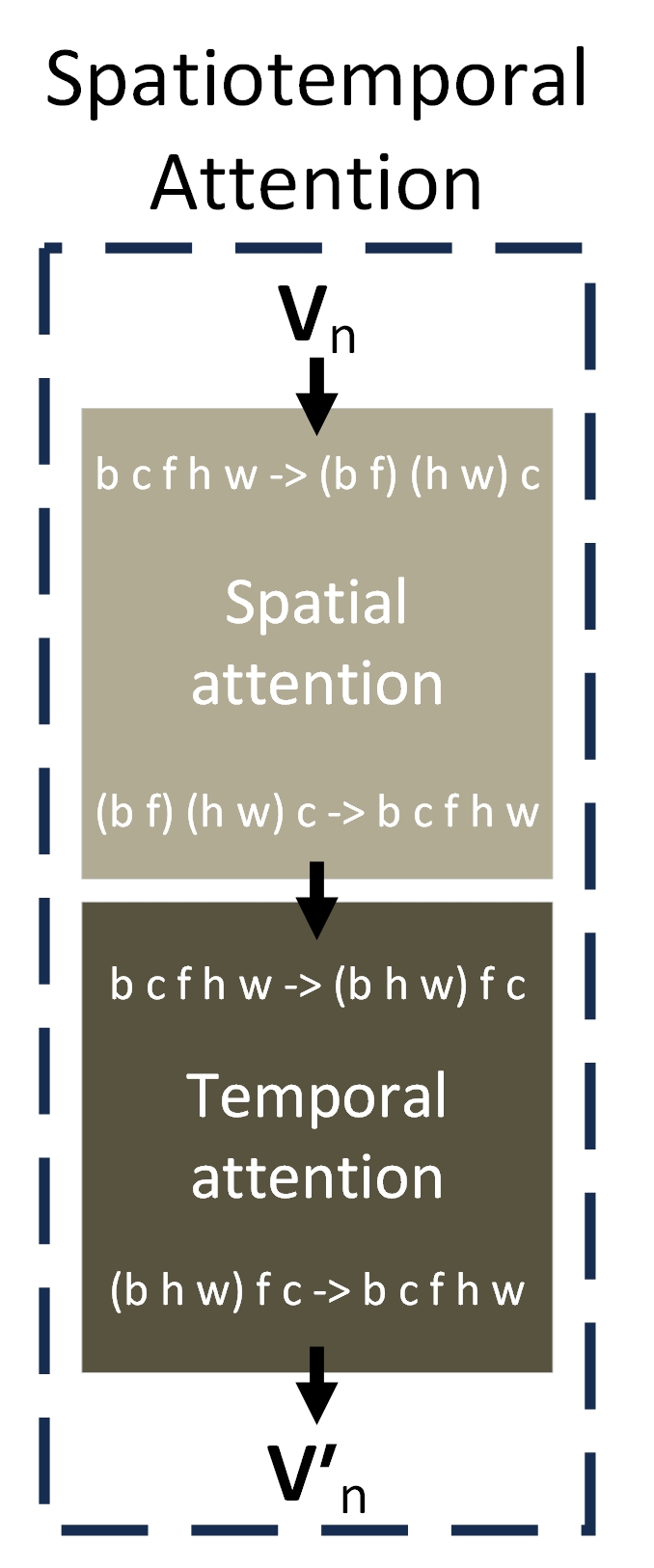}
        \caption[]%
        {}    
        \label{fig:spatiotemporal}
    \end{subfigure}
    \hfill
    \begin{subfigure}[t]{0.41\textwidth} 
        \centering
        \includegraphics[width=\columnwidth]{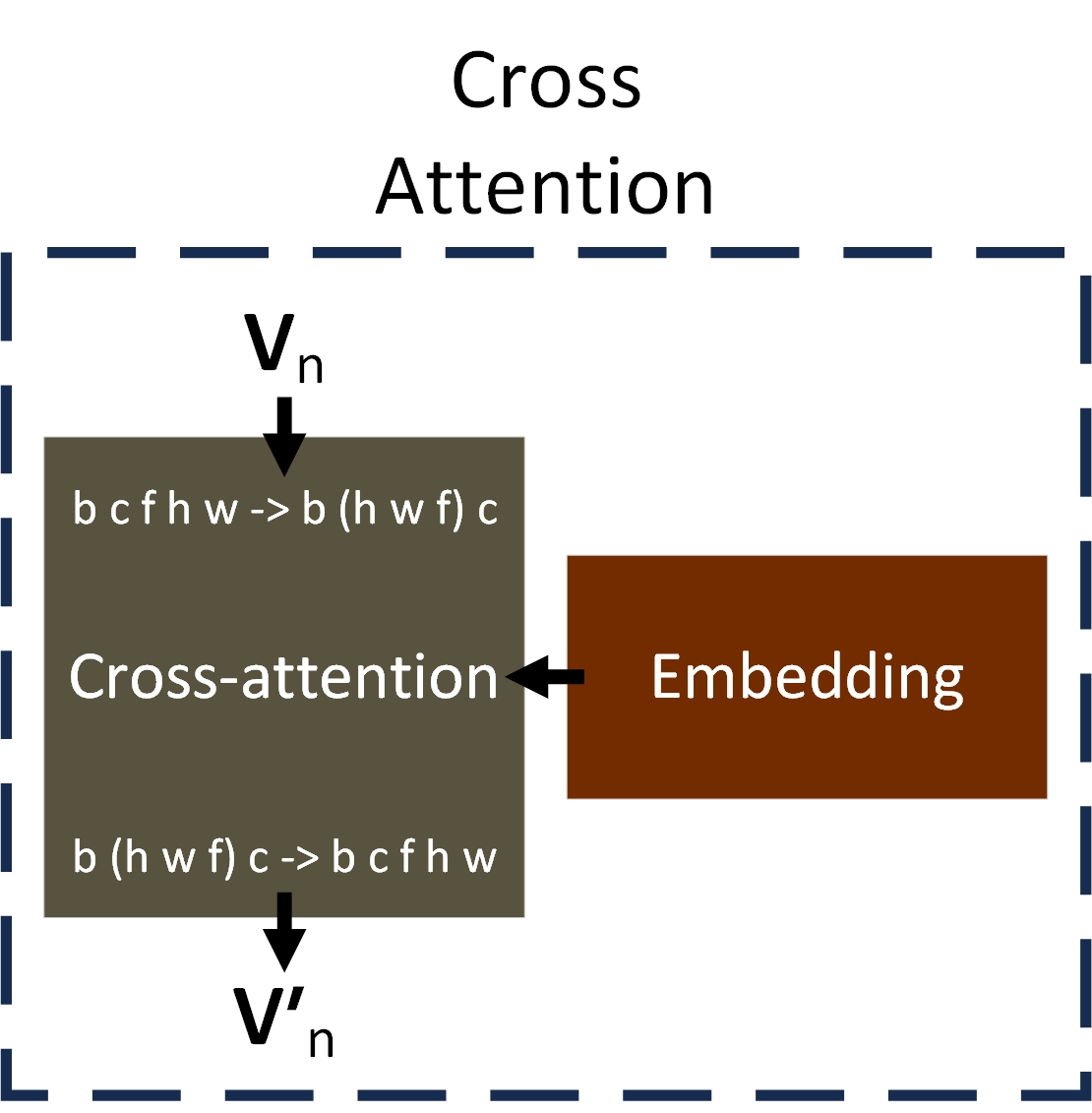}
        \caption[]%
        {}    
        \label{fig:cross}
    \end{subfigure}
    \hfill
    \caption{The architecture of the pseudo-3D convolutional and attention layers. \textit{b}, \textit{c}, \textit{f}, \textit{h}, \textit{w} represent the number or value of batch, channel, frame, height and width, respectively. (a) The pseudo-3D convolutional layer eases optimisation and performs better than its standard counterpart. (b) The spatiotemporal attention layer helps the model generate high-quality video frames while maintaining smoothness and consistency. (c) The cross-attention layer allows the model to condition the synthesised video based on the input image.}
    \label{fig:layers}
\end{figure*}

Generating high-quality videos poses significant challenges, especially when it comes to ensuring the quality and fluidity of the video are exceptional. 2D convolution layers are the most practical approach to handle the spatial aspect of an image \cite{simonyan2014very,krizhevsky2012imagenet,szegedy2015going}. While it is possible to use 3D convolution layers to generate videos, they are more difficult to optimise and can be computationally expensive, as discussed in Section~\ref{background}. We draw inspiration from the techniques proposed by \cite{chollet2017xception, singer2022make}, where they introduce an approach known as pseudo-3D convolution (shown in Fig.~\ref{fig:pseudo3D}). This method involves stacking 1D convolutional layers alongside 2D convolutional layers to efficiently process video data. Adopting the pseudo-3D convolution technique offers a more efficient and effective way to handle both spatial and temporal dimensions in video processing tasks. In Fig.~\ref{fig:pseudo3D}, we show the step-by-step transformation of $V_n$ dimensions when passing it to a pseudo-3D convolutional layer. Firstly, we change the dimension $V_n$ to (\textit{b} \textit{f}) \textit{c} \textit{h} \textit{w} for the 2D convolutional layer to process the spatial aspect. Then, we change the dimension again to (\textit{b} \textit{h} \textit{w}) \textit{c} \textit{f} for the 1D convolutional layer to process the temporal aspect.

\subsection{Frame interpolation}

The U-net decoder is trained to perform frame interpolation for video smoothing purposes. This means it adds frames between existing ones to make videos smoother and longer. We accomplish this by masking half of the frames in the latent space. The decoder then synthesises how the masked frames would look based on the neighbouring visible frames. Additionally, each frame in the latent space has four extra channels. Three of these channels depict RGB-masked video input, while the other channel is a binary channel that indicates which frames are masked.

\subsection{VAE and CLIP encoder}

We utilised the variational autoencoder (VAE) encoder in two ways - once for local conditioning and once for global conditioning. For local conditioning, we used the pretrained VAE encoder from \cite{rombach2022high} to encode the conditional image $I_c$ into $I_{vae}$. This allowed us to efficiently compress the image into a lower-dimensional latent for conditioning the video. For global conditioning, we use $I_{vae}$ and $I_{clip}$ produced by contrastive language-image pre-training (CLIP) \cite{radford2021learning}. CLIP is a model that has been pre-trained to learn visual concepts from natural language supervision. We combine $I_{vae}$ and $I_{clip}$ using an adapter proposed by \cite{karras2023dreampose}, and we use cross-attention \cite{vaswani2017attention} to condition the U-Net.

\subsection{Attention}

Attention layers are very useful for generating images and videos. They can help models prioritise and extract relevant regions of data and disregard irrelevant parts of the data \cite{guo2022attention}. There are models that have used attention for synthesising videos \cite{singer2022make, blattmann2023align}. The general equation for attention is as follows: 

\begin{equation}
    \text{Attention}(Q, K, V) = \text{softmax}\left(\frac{QK^T}{\sqrt{d_k}}\right) \cdot V
\end{equation}

where $Q$ is the query matrix representing the encoded representation of the current input, $K$ is the key matrix representing the encoded representation of the input sequence to attend to, $V$ is the value matrix containing the information to be extracted for each input element, $d_k$ is the dimensionality of the key matrixes.


In each pseudo-3D convolutional block of our model, we integrate a cross-attention layer as shown in \figurename~\ref{fig:cross}. This layer plays a crucial role in learning the significant connections between two sets of data \cite{rombach2022high}, in our case, the noisy latent video $V_n$ and the input try-on image $I_c$. The cross-attention layer calculates the relevance of each element in the noisy latent video with respect to each element in the input try-on image. By doing so, our model is able to focus on the most relevant parts from the input sources and generate the desired outcomes effectively. 

In \figurename~\ref{fig:cross}, we show that the dimension of $V_n$ has to be changed to allow the cross-attention layer to perform its task. We change the dimension of $V_n$ to \textit{b} (\textit{h} \textit{w} \textit{f}) \textit{c}. The cross-attention layer processes and modifies the texture of $V_n$ in the spatiotemporal dimension with the condition image at every channel.

The innermost layers of our U-Net use spatiotemporal attention layers that allow the model to understand complex relationships within video frames and improve quality. The spatiotemporal layers consist of spatial attention and temporal attention. Spatial attention evaluates the significance of individual elements (i.e. pixels) within a feature space relative to one another. This approach empowers the model to consider both local (neighbouring regions) and global (farther apart regions) dependencies in the data, thereby allowing it to capture intricate contextual information. Temporal attention mechanisms play a crucial role in unravelling temporal dependencies between consecutive frames within a video sequence. This enables our model to seamlessly synthesise fluid and coherent movements in video content. 

The spatiotemporal attention shown in \figurename~\ref{fig:spatiotemporal} begins with performing spatial attention on $V_n$. The dimension of $V_n$ is rearranged as (\textit{b} \textit{f}) (\textit{h} \textit{w}) \textit{c}, which enables the attention layer to process the spatial dimension at every channel. After this, the dimension is rearranged as (\textit{b} \textit{h} \textit{w}) \textit{f} \textit{c} to allow the temporal attention layer to operate and focus on the temporal dimension at every channel. Overall, the spatiotemporal attention layer helps the model to ensure that the video is high-quality and its movements look natural.  

\section{Experiment}

In this section, we will provide a comprehensive overview of the training and evaluation process for FashionFlow. Our analysis encompasses both qualitative and quantitative comparisons, offering an in-depth examination of our results. Additionally, we delve into the distinct contributions made by each model component towards the overall performance. We further elaborate on the datasets utilized in our experiments, along with a detailed exposition of the network implementation details. This commitment to transparency and thorough documentation ensures the reproducibility of our work.

\subsection{Dataset}

We trained our model on the Fashion dataset \cite{zablotskaia2019dwnet}. This dataset features professional women models who pose at various angles to showcase their dresses. There is a vast range of clothing and textures available, offering a multitude of possible appearances. 

This dataset includes 500 videos for training and 100 for testing. Each video consists of approximately 350 frames. The video resolution is set to 512 pixels in height and 400 pixels in width.

\subsection{Implementation}

We divide our proposed network into three sections: the VAE encoder, the latent U-Net, and the VAE decoder. The pre-trained VAE encoder and decoder are produced by \cite{rombach2022high}. The U-Net consists of blocks containing six pseudo-3D convolutional layers. Each block uses the same kernel sizes of 64, 128, 256, and 512. The middle section of the U-Net consists of a block with four pseudo-3D convolution layers, all utilising 512 filters. Finally, the latent space is decoded through a series of blocks containing six pseudo-3D convolutional layers. Each block includes kernel sizes of 512, 256, 128, and 64 filters, respectively. All pseudo-3D convolutional layers use a kernel size of 3, stride of 1 and padding set to 1.

After every block of pseudo-3D convolutional layers, we utilise cross-attention to enable the network to capture intrinsic detail from the conditioning image. The spatiotemporal attention is utilised in the innermost block of the U-Net.

We trained the U-Net for 2500 epochs with a denoising step of 1000. We chose DDPM \cite{ho2020denoising} as our cosine noise scheduler as it adds noise at a slower rate than linear. This enhances the diffusion model's performance \cite{nichol2021improved}. We utilised the AdamW optimiser \cite{loshchilov2017decoupled} to train the denoising U-Net. We set the learning rate hyperparameter to 0.0002 and the values of $\beta_1$ and $\beta_2$ to 0.5 and 0.999, respectively.

\subsection{Qualitative Analysis}

\begin{figure*}[!t]
    \centering
    \begin{subfigure}[t]{\textwidth}   
        \centering
        \includegraphics[width=\columnwidth]{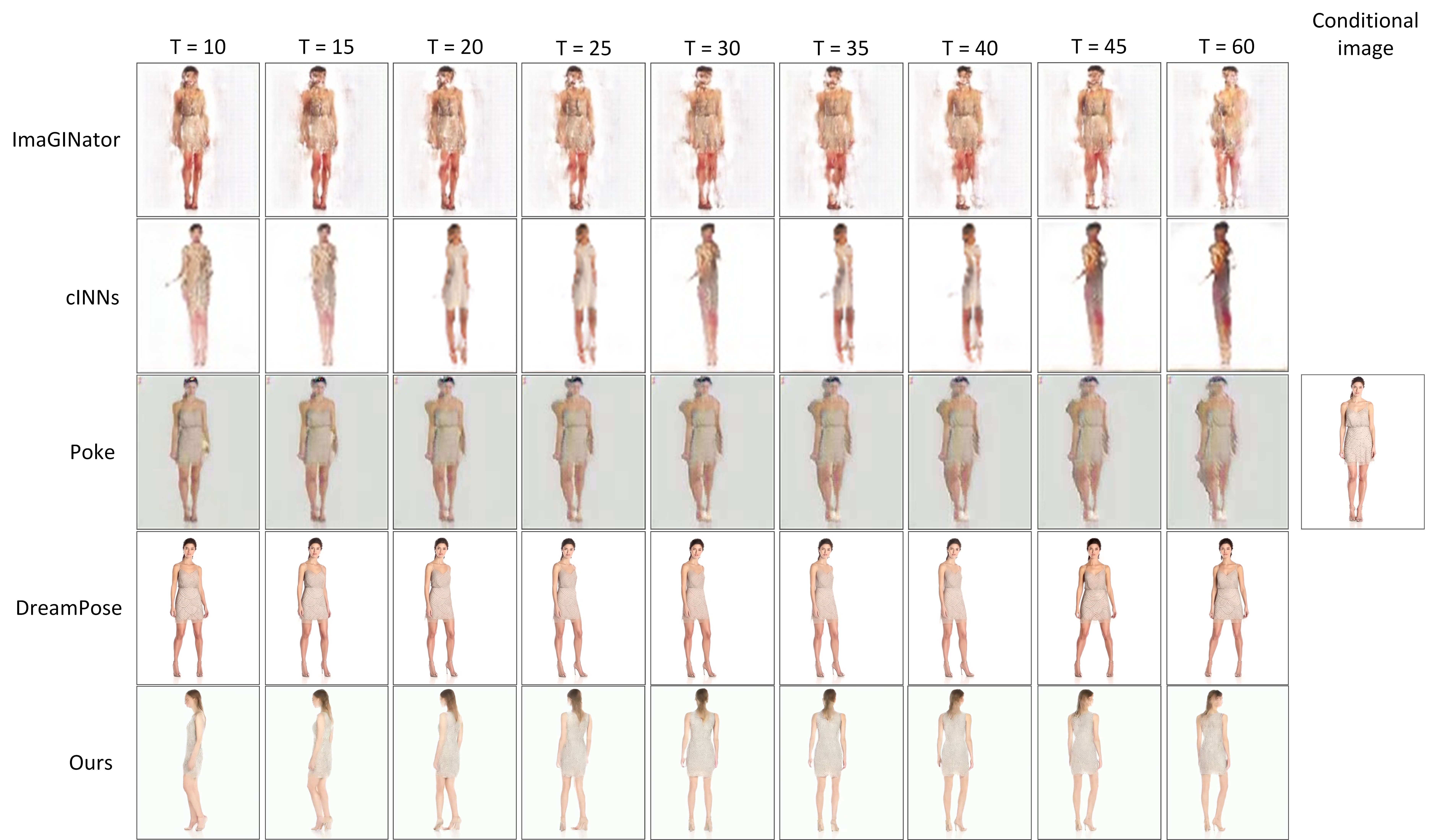}
        \caption[]%
        {} 
        \label{fig:qualitative1}
    \end{subfigure}
    \hfill
    \begin{subfigure}[t]{\textwidth} 
        \centering
        \includegraphics[width=\columnwidth]{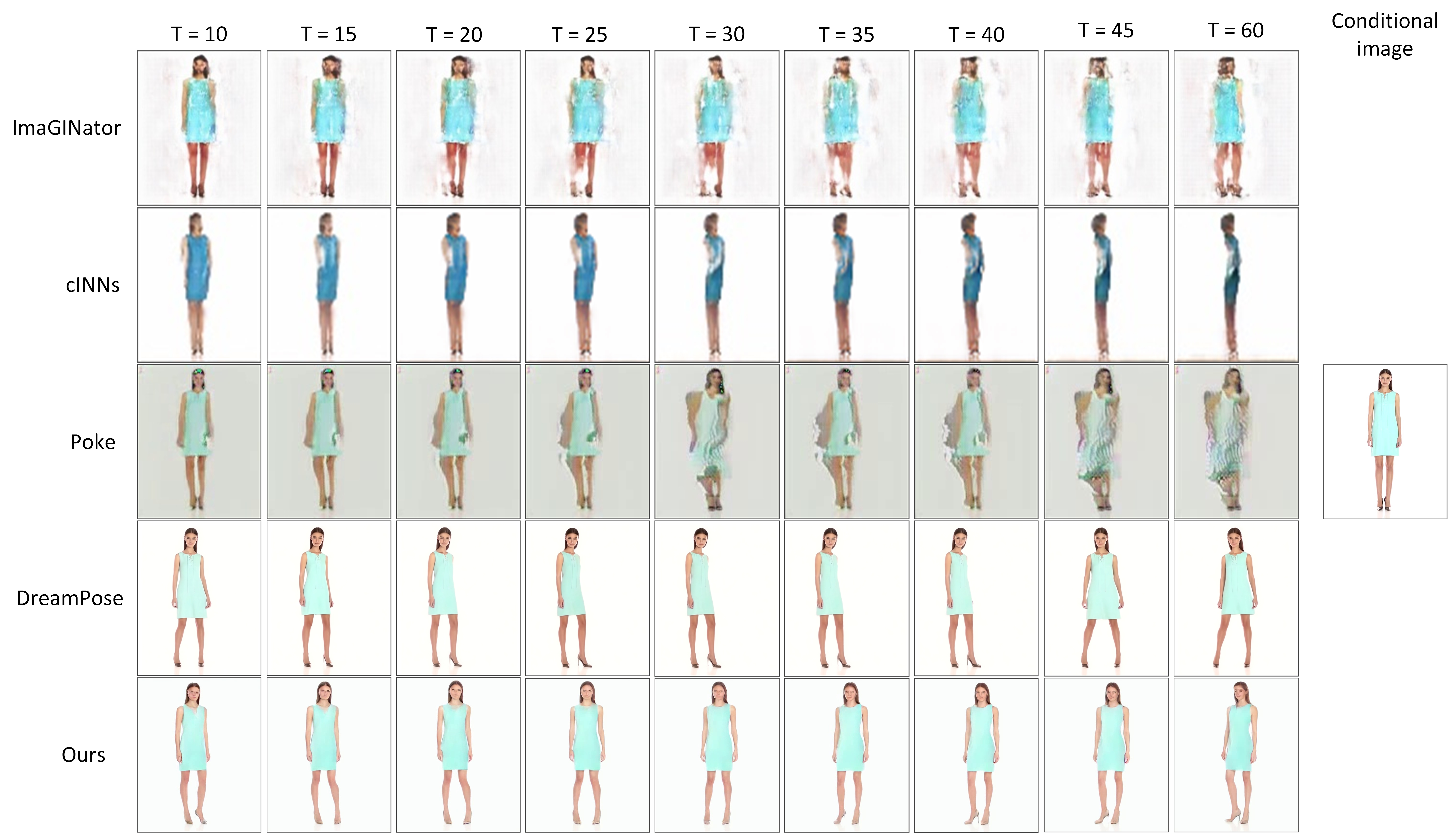}
        \caption[]%
        {}    
        \label{fig:qualitative2}
    \end{subfigure}
    \hfill
    \caption{Qualitative comparison of our method against ImaGINator \cite{wang2020imaginator}, cINNs \cite{dorkenwald2021stochastic}, Poke \cite{blattmann2021understanding} and DreamPose \cite{karras2023dreampose}. Our method performs a wider range of movements and is comparable to DreamPose in terms of quality and temporal consistency.}
    \label{fig:qualitative}
\end{figure*} 

\begin{figure*}[!t]\ContinuedFloat
    \centering
    \begin{subfigure}[t]{\textwidth} 
        \centering
        \includegraphics[width=\columnwidth]{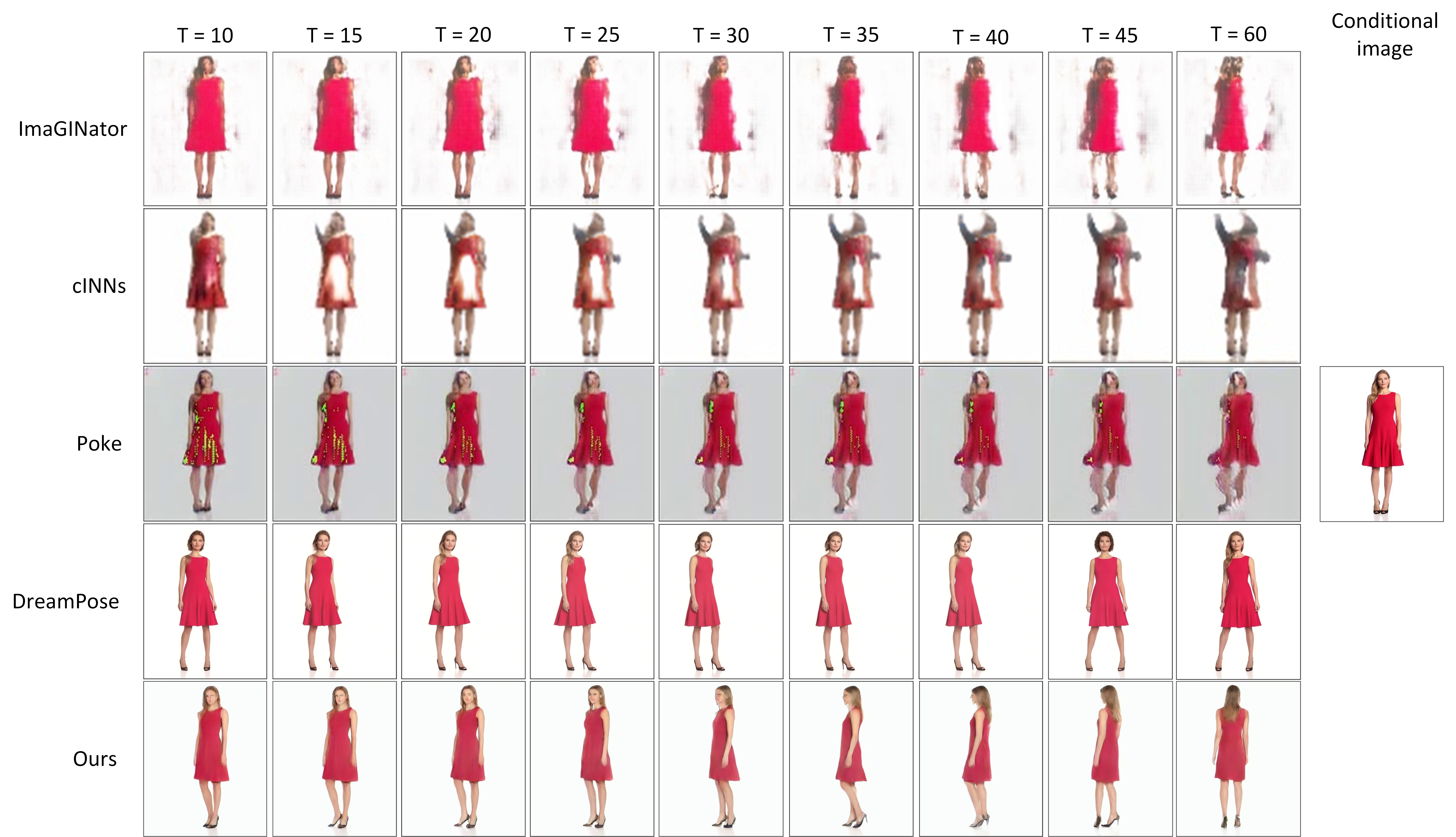}
        \caption[]%
        {}    
        \label{fig:qualitative3}
    \end{subfigure}
    \hfill
    \caption{Qualitative comparison of our method against ImaGINator \cite{wang2020imaginator}, cINNs \cite{dorkenwald2021stochastic}, Poke \cite{blattmann2021understanding} and DreamPose \cite{karras2023dreampose}. Our method performs a wider range of movements and is comparable to DreamPose in terms of quality and temporal consistency.}
    \label{fig:qualitative}
\end{figure*} 

The videos presented in \figurename~\ref{fig:qualitative} show a side-by-side comparison of our model with four other models, namely ImaGINator \cite{wang2020imaginator}, cINNs \cite{dorkenwald2021stochastic}, Poke \cite{blattmann2021understanding}, and DreamPose \cite{karras2023dreampose}. Our model outperforms others in terms of the range of motion it can perform, such as making a person turn significantly, and our result is much more temporally consistent. 

ImaGINator \cite{wang2020imaginator}, cINNs \cite{dorkenwald2021stochastic}, and Poke \cite{blattmann2021understanding} use GANs \cite{goodfellow2014generative} to synthesise videos. These models were initially designed to create short videos with a low number of frames (around 10 to 20) and a resolution of either 64x64 or 128x128. Because of this, the video quality of GAN-based models deteriorates beyond 20 frames. We used the Fashion dataset \cite{zablotskaia2019dwnet} to train ImaGINator and downloaded pre-trained models of cINNs and Poke, which were trained on the iPER dataset \cite{liu2019liquid}. The iPER dataset showcases individuals performing tai chi moves, which is similar and relevant to the movements performed in the Fashion dataset. Our approach has demonstrated better performance compared to previous work in terms of video quality, as our model generates videos with higher resolution and smoother temporal consistency. Our model generates a longer video consisting of 70 frames with a resolution of 512 for height and 640 for width. Regarding DreamPose, they were able to retain the facial details of the person better than ours because they performed person-specific fine-tuning. We do not perform this because it is significantly time-consuming and inefficient. We discuss this in Section~\ref{sec:inference_time}.

Apart from DreamPose, we encountered difficulties in accurately depicting the fine details of the face. Producing precise facial representations is a complex task. Other video diffusion models have not addressed the image-to-video problems, and there is a lack of research on how to capture intricate details such as facial features from an image. Based on the results of the text-to-video approaches, synthesising a person's face from a distance is challenging \cite{ho2022imagen, ho2022video}. Additionally, there is a lack of experiments involving human faces in the existing literature, with most research focusing on animal movements and landscape transitions. Synthesising realistic faces in videos remains a formidable challenge, necessitating further research and development efforts.

\subsection{Quantitative Analysis}

\begin{table*}[h]
\centering
\begin{tabular}{|l|l|l|l|}
\hline
Method & IS $\uparrow$ & FVD $\downarrow$ & VFID I3D  $\downarrow$ \\ \hline    
ImaGINator \cite{wang2020imaginator} & 2.003 & 3383.199 & 32.179 \\ \hline
cINNs \cite{dorkenwald2021stochastic} & 2.560 & 3325.973 & 79.386 \\ \hline
Poke \cite{blattmann2021understanding} & 2.823 & 3514.5 & 25.929 \\ \hline
Ours & \textbf{2.965} & \textbf{1659.546} & \textbf{0.867} \\ \hline
\end{tabular}
\caption{Quantitative comparison performed on the testing set of Fashion \cite{zablotskaia2019dwnet}. Our method has outperformed the GAN-based model that also generates video from images. }
\label{tab:quantitative_table}
\end{table*}

Frechet Video Distance (FVD) is an effective metric for evaluating the quality of a synthesised video \cite{unterthiner2019fvd, unterthiner2018towards}. It is influenced by Frechet Inception Distance (FID) \cite{heusel2017gans}, which is used for evaluating synthesised images. FVD introduces a feature representation that captures the temporal coherence of the content of a video, in addition to the quality of each frame. FVD consistently outperforms Structural Similarity (SSIM) and Peak Signal-to-Noise Ratio (PSNR) in terms of agreeing with human judgment \cite{unterthiner2019fvd}. FVD evaluates 16 frames of a video using the pre-trained Inflated 3D Convnet model (I3D) \cite{carreira2017quo}, which was designed to recognise the act performed in a video. The number of frames cannot be altered according to the implementation of \cite{FVD_pytorch}. However, the VFID I3D implementation by \cite{chang2019learnable} is identical to FVD, except their I3D model can evaluate up to 60 frames.

The equation for FVD and VFID I3D is denoted as:

\begin{equation}
 \begin{aligned}
d(P_R,P_G) = \lvert \mu_R + \mu_G \rvert ^2 + \mathrm{Tr}(\Sigma_R + \Sigma_G - 2(\Sigma_R \Sigma_G)^{1/2})
 \end{aligned}
\end{equation}

where $R$ represents the video from the dataset, $G$ is the generated video, the variables $\mu_R$ and $\mu_G$ represent the mean, while $\Sigma_R$ and $\Sigma_G$ represent the covariance matrices of the recorded activation data from both the real $P_R$ and generated $P_G$ videos, respectively. $P_R$ and $P_G$ were obtained by passing the videos through a pre-trained I3D, which takes into account the visual content's temporal coherence across a sequence of frames.

The Inception Score (IS) serves as a tool to assess the performance of generative models \cite{salimans2016improved}. Its purpose is to gauge both the variety and the aesthetic appeal of the images produced by these models. It achieves this by running the generated images through a classifier that has been trained beforehand and then determining a score based on the resulting probabilities. Specifically, the IS is derived from the exponential of the average KL divergence between the class distribution of the generated images and the class distribution of a large collection of real images. A higher Inception Score suggests that the generated images possess greater diversity and visual appeal.

\begin{equation}
\text{IS} = \exp\left(\mathbb{E}_{x\sim P_G}\left[D_{\text{KL}}(P(y|x) \| P(y))\right]\right)
\label{eq:IS}   
\end{equation}

where $P_G$ represents the distribution of the generated images, the notation $\mathbb{E}_{x\sim P_G}$ signifies that we are taking the average over samples $x$ drawn from this distribution. The conditional distribution $P(y|x)$ indicates how labels $y$ (values obtained from a pre-trained classifier) are distributed when we have a generated image $x$. The marginal distribution $P(y)$, on the other hand, shows the overall label distribution. The Kullback-Leibler divergence, denoted as $D_{\text{KL}}$, quantifies how dissimilar these two distributions are. The formula computes the expected value, denoted as $\mathbb{E}$, of the Kullback-Leibler divergence across all the generated images. Finally, the exponential function $\exp$ is applied to this expected Kullback-Leibler divergence to yield the IS. 

\tablename~\ref{tab:quantitative_table} showcases the results of our method in comparison to three other models, namely ImaGINator \cite{wang2020imaginator}, cINNs \cite{dorkenwald2021stochastic}, and Poke \cite{blattmann2021understanding}. The comparison was made on the Fashion dataset's testing subset \cite{zablotskaia2019dwnet}. Although we downloaded the pretrained models of cINNs and Poke, trained on the iPER dataset \cite{liu2019liquid}, the movements performed are similar to those in the Fashion dataset \cite{zablotskaia2019dwnet}, such as turning and stepping side to side. Our model has outperformed the previous works in all metrics. The results were particularly impressive for the metrics FVD and VFID I3D. We did not conduct a quantitative evaluation on DreamPose \cite{karras2023dreampose} because fine-tuning person-specific models of 100 people is impractical and time-consuming. Refer to Section~\ref{sec:inference_time} for explanation. 

\subsection{Inference Time}
\label{sec:inference_time}

Our model can synthesise a video consisting of 70 frames 9.3x faster than DreamPose. This speed comparison does not include the additional time required to fine-tune DreamPose's U-Net and VAE decoder to generate a person-specific model. If we were to factor in the additional time required for fine-tuning DreamPose, it would take even longer to synthesise a video. Our model is universal, which means it only needs a conditioning image, and no fine-tuning is necessary.

Generating videos is a complex task and requires a practical and efficient generative model. Fine-tuning a model for each individual is highly impractical as it takes a considerable amount of time and requires a large amount of additional storage space to save weight for each person. Our model is a more practical and faster alternative that does not require fine-tuning, nor does it consume additional storage space. Additionally, we believe that users would be highly frustrated if they had to wait for an AI model to fine-tune and then synthesise the video. In a business context, where multiple customers are being served, this would create a huge queue for those customers who want to use the model. Therefore, speed and efficiency are crucial factors when serving multiple people. 

\subsection{Ablation Study}

\begin{table*}[h]
\centering
\begin{tabular}{|l|l|l|l|}
\hline
Method & IS $\uparrow$ & FVD $\downarrow$ & VFID I3D $\downarrow$ \\ \hline
Global only & 2.891 & 1812.462 & 0.881 \\ \hline      
Local only & 2.801 & 2111.037 & 1.179 \\ \hline
Global and local & 2.965 & 1659.546 & 0.867 \\ \hline
\end{tabular}
\caption{Ablation study on the conditioning methods. Our model performs best when both global and local conditioners are employed.}
\label{tab:ablation_table}
\end{table*}

In this section, we will conduct an ablation study to examine the effectiveness of global and local conditioning. Global conditioning refers to using cross-attention mechanisms to influence the entire U-Net architecture. Meanwhile, local conditioning involves inserting the encoded image as the first frame of our noise input.

\begin{figure*}
\centering
  \includegraphics[width=\columnwidth]{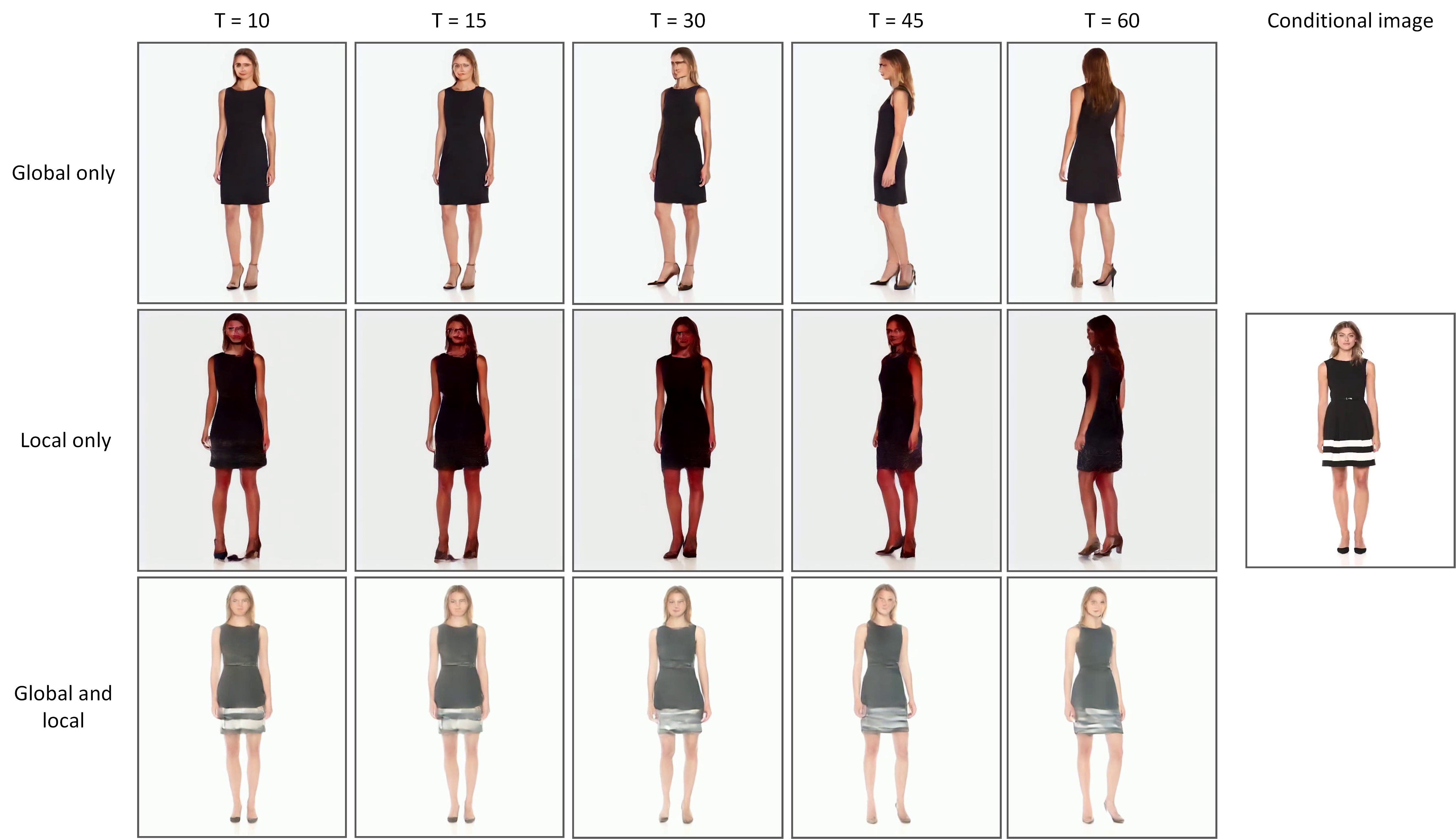}
  \caption{The effects of image conditioning. Global conditioning captures the overall colour of the garment, but it misses out on smaller details like the white stripes. Local conditioning darkened the skin colour too much and also failed to capture small clothing details. Using both local and global conditioning, it captures the overall colour from the conditioning image, and the model was able to pick up small details like the stripe.}
  \label{fig:ablation}
\end{figure*}

In \figurename~\ref{fig:ablation}, we can observe the effects of conditioning image generation models based on local and global factors. The first row of the figure shows the results of using global conditioning. We can see that the model fails to capture small details, such as the white stripe on the dress. Instead, it captured the overall colour, which is inaccurate. The second row shows the results of using only local conditioning. The model also failed to capture fine-grained detail, and as the video progresses, the colour is altered since the model did not capture enough information from the first frame. Finally, the third row shows the results of using both global and local conditioning, which yields the best result. In this case, the model does a better job of preserving specific details, such as the white stripe on the dress, while also maintaining the overall colour scheme. This is also supported by \tablename~\ref{tab:ablation_table}. Using both conditioners has quantitatively outperformed methods using a single variant across all metrics.  

\subsection{Limitations}

\begin{figure*}
\centering
  \includegraphics[width=\columnwidth]{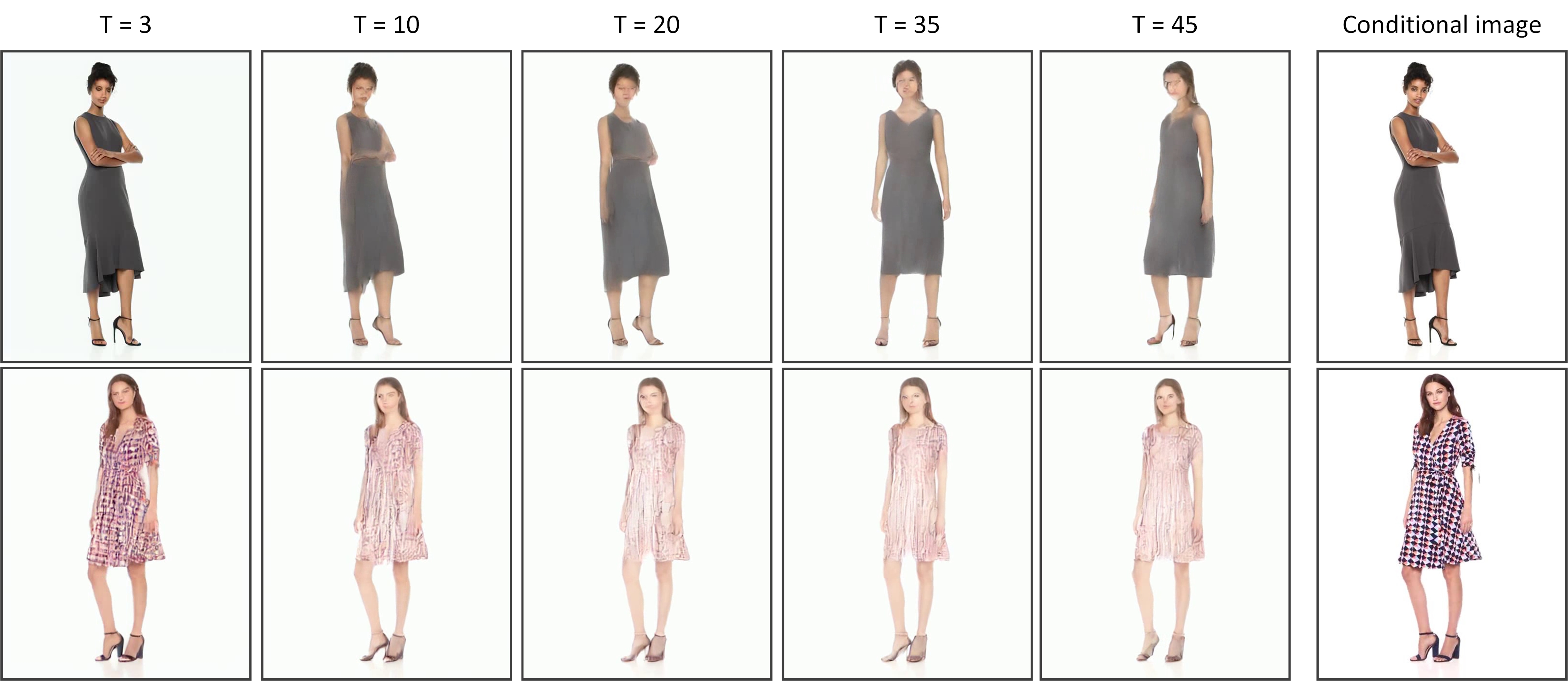}
  \caption{Limitations of our model. In the first row, our model fails to accurately preserve the skin colour of ethnic minorities because the Fashion dataset \cite{zablotskaia2019dwnet} underrepresents them. Our model produces skin colours that are most common in the dataset, as it hasn't properly learned how to preserve skin colour. The second row shows that our model struggles with clothes that have complex patterns. The patterns are corrupted in the video, which may be due to the global conditioner not passing enough information to the generator.}
  \label{fig:limitations}
\end{figure*}

Currently, our model has a few areas that require improvement. The videos generated by our model sometimes have colours that look faded, facial identities that are poorly preserved, and clothing details that are lost as the video progresses. The second row of \figurename~\ref{fig:limitations} illustrates this. We believe that the global conditioning in the U-Net model is hindering its ability to capture intricate details from the conditioning image. This causes the U-Net to lose vital information about the conditioning image, leading to more hallucinations and a reduced degree of fidelity. We leave the improvement of information flow from the global conditioner for future research.

The Fashion dataset \cite{zablotskaia2019dwnet} has some limitations. One of the main limitations is that it is not representative of the entire population. This dataset only focuses on dresses and does not include models of men, children, or ethnic minorities. The first row of \figurename~\ref{fig:limitations} shows that the absence of representative data can result in a lower degree of fidelity in the generated video, where the skin colour is not preserved. Instead, it has been generated to match the most common skin tones present in the dataset. Due to this limitation, it may not be very useful for businesses that want to cater to a wider demographic. In order for our model to be more useful, there needs to be a dataset that is diverse.

\subsection{Conclusion}

Our research introduces a novel architecture for diffusion models that is tailored specifically for synthesising high-fidelity fashion videos using conditional images. We propose a methodology that utilises pseudo-3D convolution, VAE, and CLIP encoder to condition synthesised videos both on a global and local scale. Our approach represents a significant advancement over previous efforts in this domain, such as synthesising videos at a faster pace and not requiring person-specific fine-tuning. Additionally, it produces videos that are temporally coherent and capture vital details from the conditioning image.

We conducted a thorough comparison of our image-to-video model with various other models. We demonstrated that the video quality produced by our model is significantly better than the GAN-based methods. Our model generates videos with a higher resolution, allowing for a more detailed output. Additionally, we observed that our model produces videos that have better temporal consistency, meaning that the frames flow more seamlessly from one to the other.

We conducted an ablation study to evaluate the effectiveness of our approach. Our results show that conditioning the U-Net on local and global scales allows the model to preserve the most detail from the condition image. By doing so, our method can generate high-quality videos that demonstrate a high degree of fidelity and realism.

Overall, our work highlights the potential of combining deep learning techniques to synthesise high-quality videos with greater efficiency and precision. Our approach has significant implications for the fashion industry, where the ability to create high-quality videos is becoming increasingly important for marketing and showcasing products in a competitive setting.

\printbibliography
\end{document}